\documentclass[conference]{IEEEtran}
\IEEEoverridecommandlockouts
\usepackage{cite}
\usepackage{amsmath,amssymb,amsfonts}
\usepackage{algorithmic}
\usepackage{graphicx}
\usepackage{textcomp}
\usepackage{xcolor}
\def\BibTeX{{\rm B\kern-.05em{\sc i\kern-.025em b}\kern-.08em
    T\kern-.1667em\lower.7ex\hbox{E}\kern-.125emX}}
\begin{document}

\title{Mapping New Realities: Ground Truth Image Creation with Pix2Pix Image-to-Image Translation}

\author{\IEEEauthorblockN{1\textsuperscript{st} Zhenglin Li}
\IEEEauthorblockA{\textit{Department of Computer Science and Engineering} \\
\textit{Texas A\&M University}\\
College Station, TX, USA \\
zhenglin\_li@tamu.edu}
\and
\IEEEauthorblockN{2\textsuperscript{nd} Bo Guan}
\IEEEauthorblockA{\textit{College of Science} \\
\textit{Virginia Tech}\\
Blacksburg, VA, USA\\
jasonguan0107@gmail.com}
\and
\IEEEauthorblockN{4\textsuperscript{th} Yiming Zhou}
\IEEEauthorblockA{\textit{Department of Engineer Sciences} \\
\textit{Saarland University of Applied Sciences}\\
Saarbruecken, Germany \\
yiming.zhou@htwsaar.de}
\and
\IEEEauthorblockN{3\textsuperscript{rd} Yuanzhou Wei}
\IEEEauthorblockA{\textit{Department of Electrical and Computer Engineering} \\
\textit{Florida International University}\\
Miami, FL, USA \\
ywei011@fiu.edu}
\and
\IEEEauthorblockN{5\textsuperscript{th} Jingyu Zhang}
\IEEEauthorblockA{\textit{Division of the Physical Sciences} \\
\textit{The University of Chicago}\\
Chicago, IL, USA \\
simonajue@gmail.com}
\and
\IEEEauthorblockN{6\textsuperscript{th} Jinxin Xu}
\IEEEauthorblockA{\textit{Department of Cox Business School} \\
\textit{Southern Methodist University}\\
Dallas, TX, USA \\
jensenjxx@gmail.com}
}

\maketitle

\begin{abstract}
Generative Adversarial Networks (GANs) have significantly advanced image processing, with Pix2Pix being a notable framework for image-to-image translation. This paper explores a novel application of Pix2Pix to transform abstract map images into realistic ground truth images, addressing the scarcity of such images crucial for domains like urban planning and autonomous vehicle training. We detail the Pix2Pix model's utilization for generating high-fidelity datasets, supported by a dataset of paired map and aerial images, and enhanced by a tailored training regimen. The results demonstrate the model's capability to accurately render complex urban features, establishing its efficacy and potential for broad real-world applications.
\end{abstract}


\section{Introduction}
The advent of Generative Adversarial Networks (GANs) has revolutionized the field of image processing, enabling new heights of image synthesis and transformation \cite{creswell2018generative, liao2022self, liao4583223self, yin2022learning}. Among these advancements, the Pix2Pix framework has emerged as a pivotal methodology for image-to-image translation tasks, particularly in applications that require a high degree of accuracy and detail, such as the construction of ground truth images from cartographic data. This paper introduces a novel application of Pix2Pix, leveraging its conditional adversarial network structure to transform abstract map images into detailed, realistic ground truth images, which has a lot of real-world applications \cite{zhan2021deepmtl}. 
  
This research is rooted in the premise that accurate ground truth images are indispensable across numerous domains, from urban planning and simulation to autonomous vehicle training and geographical information systems. However, the scarcity of such images often poses a significant barrier. By employing the Pix2Pix model, we propose a scalable solution to this scarcity, with the potential to automate and enhance the generation of high-fidelity ground truth datasets. 

\section{Related Study}

The challenge of synthesizing photorealistic images from schematic representations is multifaceted, involving not only the faithful reproduction of spatial information but also the nuanced translation of textures, patterns, and contextual details. Pix2Pix addresses this challenge by conditioning the generation process on input images, herein maps, and guiding the generative model with a discriminator that assesses the verisimilitude of the output against the actual ground truth. The discriminator's role is crucial, as it pushes the generator towards creating outputs that are increasingly indistinguishable from real-world imagery. 

Convolutional Neural Networks (CNNs) have been instrumental in advancing image processing tasks, providing a foundational architecture for many generative models \cite{wang2019wear, ding2018vehicle}. Recently, image to image translation presents a promising result. Among those popular image to image translation models, Adversarial Training play a significant role. 

\section{Methodology }

\subsection{Data Preparation}

In the initial stage of our research, we compiled a dataset comprising pairs of images: one representing a map and the other its corresponding real-world aerial view, the ground truth. These images were processed using TensorFlow's I/O and image decoding functions to load and pair JPEG images accurately. The images were split and cast into floating-point tensors to facilitate further operations \cite{wang2024balanced, li2022domain}. 

\subsection{Image Preprocessing}

To enhance the robustness of our model, we applied random jittering and normalization to the input data. The random jittering process involves resizing the images to 286x286 pixels and cropping them to the final size of 256x256 pixels \cite{zhu2021taming, lai2023detect}. Furthermore, we randomly mirror the images on the horizontal axis. The normalization step adjusts the pixel values to a [-1, 1] range, which is a standard practice aimed at promoting training convergence \cite{sun2022rethinking, jia2023kg, li2024cpseg}. 

\subsection{Model Architecture} 
Our model architecture is based on the Pix2Pix framework, which consists of a generator and a discriminator  \cite{isola2017image}. The generator aims to transform an input image into an output image, while the discriminator tries to distinguish between the real target image and the generated image.

\subsection{Generator  } 
\begin{figure}
    \centering
    \includegraphics[width=1\linewidth]{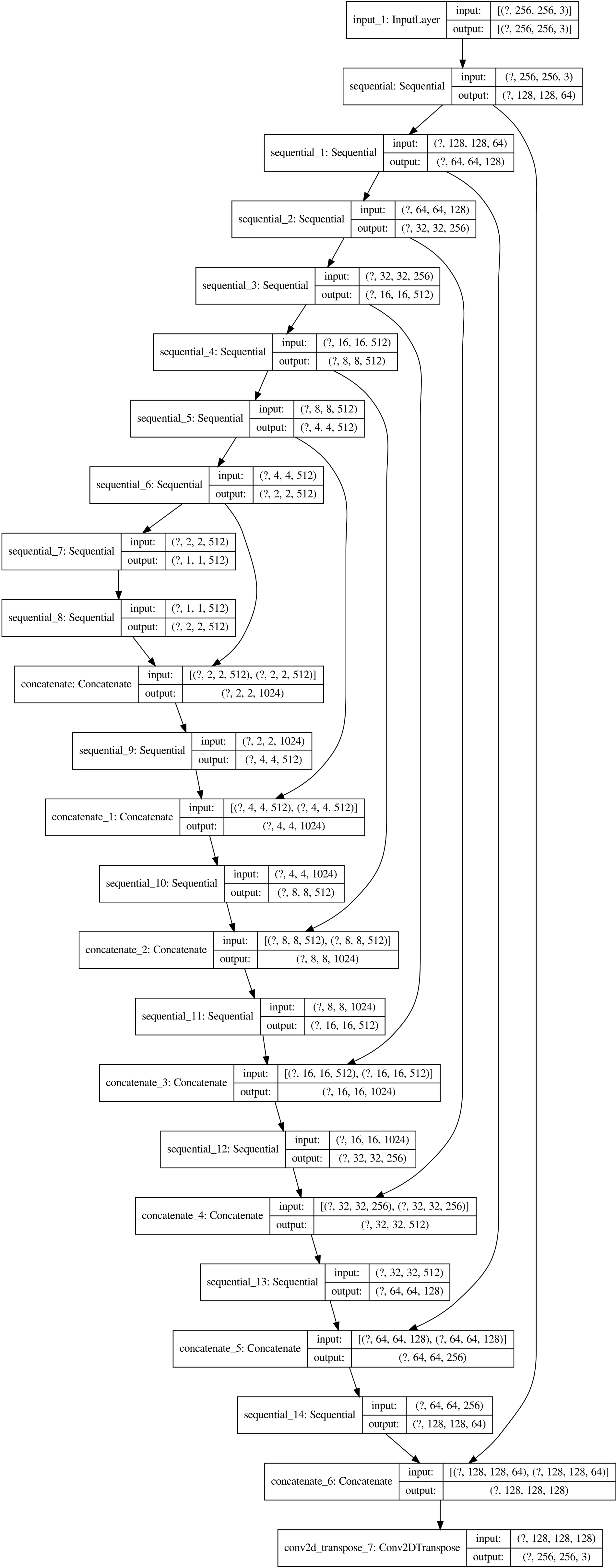}
    \caption{The Architecture of Generator. It is a U-Net-like architecture with an encoder-decoder structure.}
    \label{fig1}
\end{figure}

The generator is a U-Net-like architecture with an encoder-decoder structure \cite{zhuang2019lighter}.  Fig.~\ref{fig1} shows such architecture. 

The encoder consists of a series of convolutional layers with downsampling, and the decoder consists of a series of transposed convolutional layers with upsampling. Skip connections are used between corresponding layers in the encoder and decoder to help preserve spatial information \cite{zhou2024pass}.

The encoder consists of the following layers:
\begin{itemize}
    \item Convolutional layer with 64 filters, kernel size of 4, stride of 2, and no batch normalization.
    \item Convolutional layers with 128, 256, 512, 512, 512, 512, and 512 filters, each with a kernel size of 4, stride of 2, and batch normalization.
\end{itemize}

The decoder consists of the following layers:
\begin{itemize}
    \item Transposed convolutional layers with 512 filters, kernel size of 4, stride of 2, batch normalization, and dropout.
    \item Transposed convolutional layers with 512, 512, 512, 256, 128, and 64 filters, each with a kernel size of 4, stride of 2, and batch normalization.
    \item Transposed convolutional layer with 3 output channels (for RGB images), kernel size of 4, stride of 2, and a tanh activation function.
\end{itemize}

The output of the generator, \(G(x)\), for an input image \(x\) is given by: 

\begin{equation}
\begin{aligned}
G(x) &= \tanh \left(W_n \ast \ldots \sigma \left(W_2 \ast \sigma \left(W_1 \ast x + b_1\right) \right. \right. \\
     &\quad \left. \left. + b_2\right) \ldots + b_n\right)
\end{aligned}
\end{equation}

where \(W_i\) and \(b_i\) are the weights and biases of the \(i\)-th layer and \(\sigma \) is the activation function.

\subsection{Discriminator    } 
\begin{figure}
    \centering
    \includegraphics[width=1\linewidth]{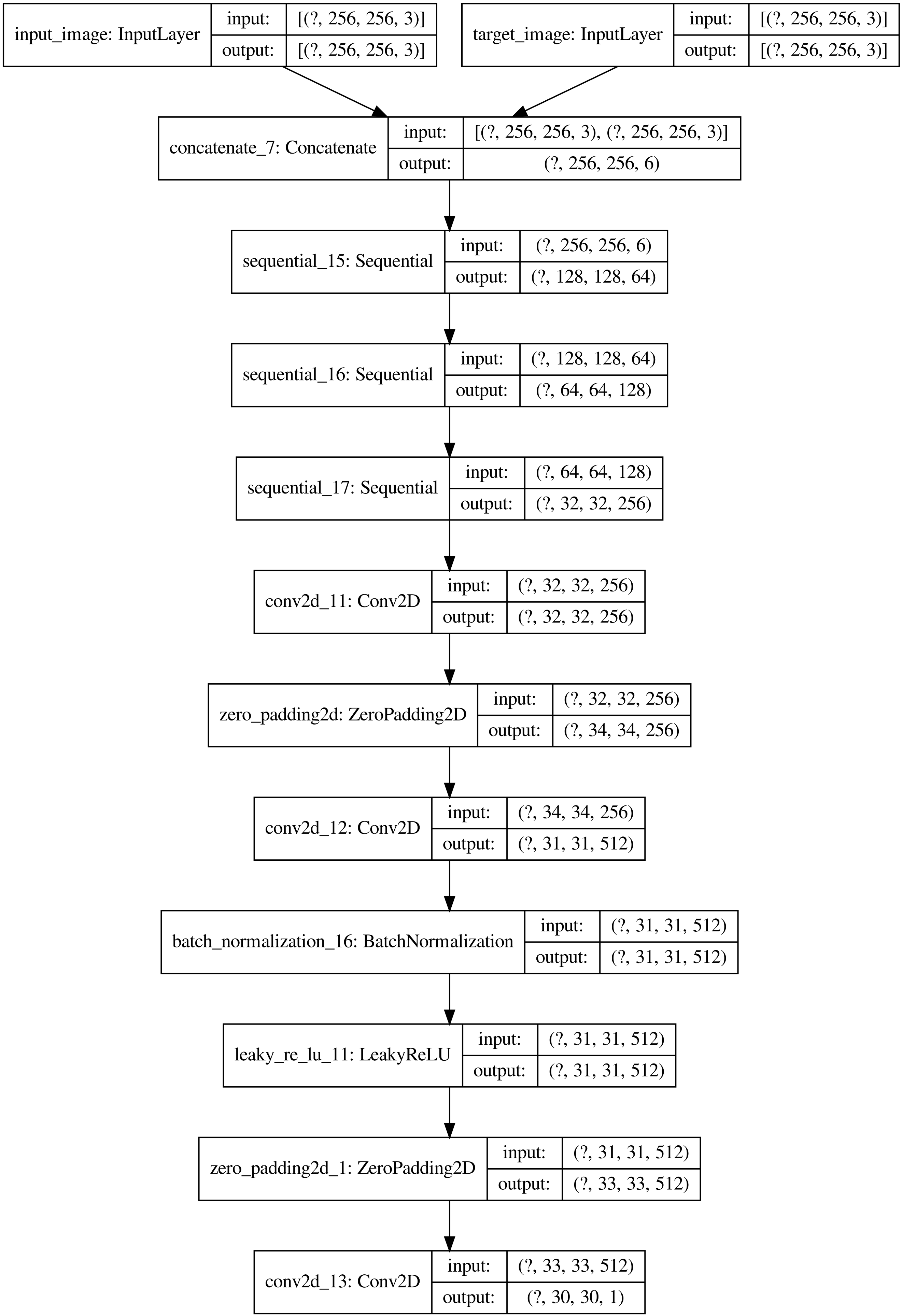}
    \caption{The Architecture of Discriminator}
    \label{fig2}
\end{figure}

The discriminator is a PatchGAN classifier, tailored for tasks involving paired input and target images to do the image translation. It consists of input layers, concatenation layer, three downsampling layers, convolutional layers, zeo padding layers, batch normalization layer, activation layers using LeakyReLU, and final convolutional layer. Fig.~\ref{fig2} shows such architecture.

\begin{itemize}
    \item Input Layers: Two input layers for the input image and the target image, each with a shape of [256, 256, 3].

    \item Concatenation Layer: Combines the input and target images along the channel dimension, resulting in a shape of [256, 256, 6].

    \item Downsampling Layers: First downsample reduces dimensions to [128, 128, 64], with 64 filters and a kernel size of 4. Second downsample further reduces dimensions to [64, 64, 128], with 128 filters. Third downsample reduces dimensions to [32, 32, 256], with 256 filters.

    \item Convolutional Layers:  A convolutional layer with 256 filters and kernel size of 3, activation function relu and another convolutional layer with 512 filters, kernel size of 4, and strides of 1, without bias units, initiated with a specific kernel initializer.

    \item Zero Padding Layers: One ZeroPadding2D applied after the first convolutional layer (256 filters) to adjust the spatial dimensions before further processing. And another ZeroPadding2D applied after the LeakyReLU activation, prior to the final convolutional operation.

    \item Batch Normalization Layer: BatchNormalization applied after the second convolutional layer (512 filters) to normalize the activations of the previous layer.

    \item Activation Layers: LeakyReLU provides a non-linear activation function that allows a small gradient when the unit is not active.

    \item Final Convolutional Layer: A final convolutional operation with 1 filter, kernel size of 4, and strides of 1 to produce a single-channel output.

\end{itemize}

The discriminator takes as input the concatenation of the input image and the target/generated image and outputs a single scalar value representing the probability that the input image-target pair is real. Both the generator and discriminator use Leaky ReLU activation functions and are initialized with random normal weights with a mean of 0 and a standard deviation of 0.02. 

The output of the discriminator, \(D(x, y)\), for an input image \(x\) and target image \(y\) is given by:

\begin{equation}
\begin{aligned}
D(x,y) &= \sigma \Bigg( W_m \ast \ldots \ast \sigma \Big( W_2 \ast \sigma \left( W_1 \ast \begin{bmatrix} x \\ y \end{bmatrix} + b_1 \right) + b_2 \Big) \\
&\quad \ldots + b_m \Bigg)
\end{aligned}
\end{equation}

where [\(x\),\(y\)] denotes the concatenation of \(x\) and \(y\), and \(W_i\) and \(b_i\) are the weights and biases of the \(i\)-th layer.

\section{Experiment and Result }

During the experiment, we evaluated the performance of the Pix2Pix model in translating abstract map representations into corresponding ground truth images. The model was trained on a dataset comprising various urban and rural landscapes to ensure robustness and generality in the translation capability \cite{wang2022interactive}.

\subsection{Training Procedure    } 
The training is conducted iteratively, alternating updates between the generator and discriminator using TensorFlow's gradient computation and application functionalities. This process is repeated across several epochs, with periodic generation of images to visually monitor translation quality \cite{sun2020substituting}.

\subsection{Optimization Strategy  } 
We employ the Adam optimizer for both generator and discriminator, with a learning rate of 2e-4 and a momentum parameter \(\beta_1\) set to 0.5. This choice is made to facilitate stable training and convergence.



\subsection{Result Visualization  } 

\begin{figure}
    \centering
    \includegraphics[width=1\linewidth]{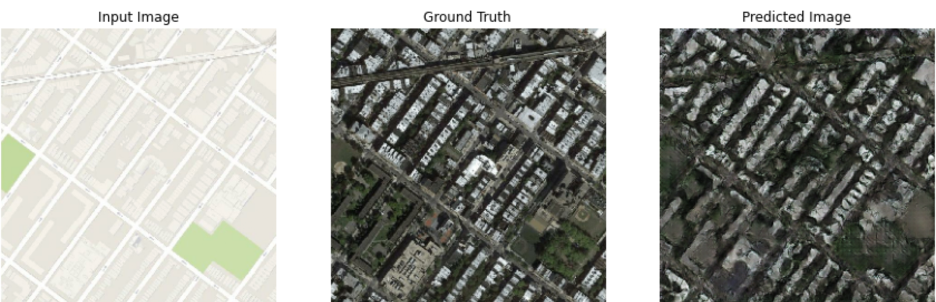}
    \includegraphics[width=1\linewidth]{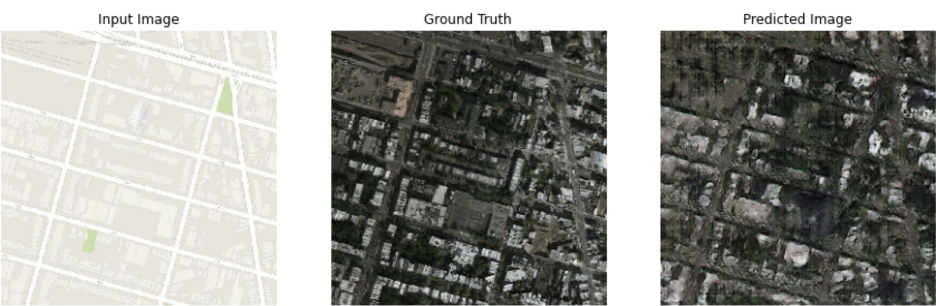}
    \includegraphics[width=1\linewidth]{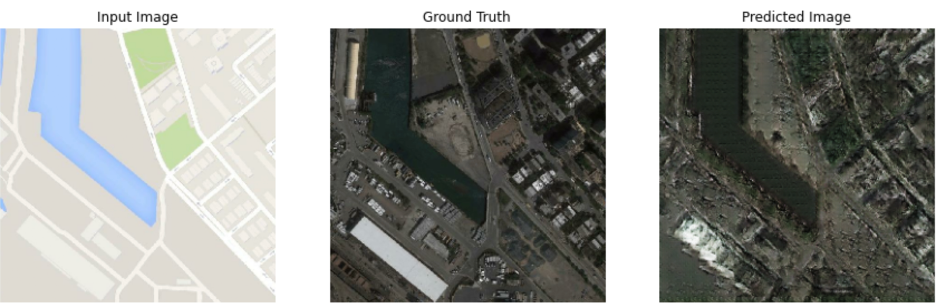}
    \caption{Result Visualization}
    \label{fig3}
\end{figure}

Furthermore, we present a series of figure comparisons to illustrate the model's translation capability. Fig.~\ref{fig3} showcase the three sets of input maps, ground truth, and Pix2Pix-generated images. These visual results highlight the model's proficiency in generating coherent and contextually accurate urban imagery.

\section{Limitations and Future Work }
 
The results from our experiments with the Pix2Pix model confirm its potential as a powerful tool for map-to-image translation tasks. The ability to generate detailed and accurate ground truth images from abstract representations holds significant promise for various applications, including urban planning, simulation environments, and geospatial analysis. 

Despite the promising results, the model exhibited limitations in certain scenarios. Specifically, regions with homogeneous textures or repetitive patterns occasionally resulted in artifacts in the generated images. 

Future work may explore the integration of more sophisticated loss functions and training strategies to enhance the model's performance and reduce the incidence of translation artifacts. Further investigations could delve into the utilization of advanced loss functions that are specifically tailored to address the unique challenges of the domain, thereby potentially yielding higher quality translations \cite{zhang2023unleashing}. Additionally, exploring a variety of training strategies, such as adaptive learning rate schedules or the use of transfer learning  \cite{wang2023transfer}, could significantly improve the model's ability to generalize across different contexts. Beyond traditional methods, the incorporation of transformer-based models \cite{lin2023comprehensive, li2023bubble, liu2023data}, multi-modal approaches \cite{lyu2022multimodal, lin2023mmst}, vision-language integrations \cite{zhou2024visual, chen2024taskclip, xin2024parameter, chen2023generative}, and reinforcement learning techniques could offer promising avenues for research . These methods may not only refine the accuracy but also broaden the scope of the model’s applicability to more complex scenarios, enhancing its robustness and effectiveness.
 

\section{Conclusion}
In this study, we have demonstrated the effective application of the Pix2Pix model to transform abstract map images into highly detailed and realistic ground truth images, suitable for use in urban planning and autonomous vehicle training. The Pix2Pix framework, employing a conditional adversarial network, has shown remarkable capability in accurately rendering complex urban landscapes and features from simple cartographic inputs.

Our experiments highlight the model's robustness and its potential to significantly alleviate the scarcity of high-fidelity ground truth datasets, a critical barrier in numerous domains. By leveraging a dataset of paired map and aerial images and optimizing our training procedures, we achieved impressive results that validate the practical utility of the Pix2Pix model.

Overall, our findings support the Pix2Pix model's potential as a transformative tool for image-to-image translation tasks, opening avenues for its application in diverse fields that rely on accurate and detailed visual data representations.

\bibliographystyle{ieeetr}
\bibliography{main}

\end{document}